\pdfoutput=1

\documentclass[11pt]{article}

\usepackage{acl}

\usepackage{times}
\usepackage{latexsym}

\usepackage[T1]{fontenc}

\usepackage[utf8]{inputenc}

\usepackage{microtype}
\usepackage{graphicx}
\usepackage{eurosym}
\usepackage{tablefootnote}
\graphicspath{ {images/} }
\usepackage{hyperref}
\usepackage{textcomp}
\usepackage{hhline}
\usepackage[T1]{fontenc}
\usepackage{listings}
\usepackage{stfloats}
\usepackage{csquotes}
\usepackage{array}
\usepackage{booktabs}

\usepackage[super]{nth}
\usepackage[inline,shortlabels]{enumitem}
\usepackage{multirow}
\usepackage{makecell}
\usepackage{siunitx}
\usepackage{multicol}

\usepackage{svg}
\usepackage{float}
\usepackage{ amssymb }
\usepackage{amsmath,amsfonts,amssymb}

\usepackage{graphicx}

\usepackage{amsmath,amsfonts,amssymb}
\graphicspath{ {images/} }

\title{Efficient Hierarchical Domain Adaptation \\ for Pretrained Language Models}

\author{ 
	Alexandra Chronopoulou$^{1}$\thanks{ \hspace{1mm} Work done while an intern at AllenAI.} ,
	Matthew E. Peters$^{2}$,
    Jesse Dodge$^{2}$\\
	$^1$Center for Information and Language Processing, LMU Munich, Germany \\
	$^2$Allen Institute for Artificial Intelligence, Seattle, WA \\

	{ \tt achron@cis.lmu.de} 
	\\{\tt \{matthewp,jessed\}@allenai.org
	}}
\begin{document}
\maketitle

\begin{abstract}
The remarkable success of large language models has been driven by dense models trained on massive unlabeled, unstructured corpora. These corpora typically contain text from diverse, heterogeneous sources, but information about the source of the text is rarely used during training. Transferring their knowledge to a target domain is typically done by continuing training in-domain. 
In this paper, we introduce a method to permit domain adaptation to many diverse domains using a computationally efficient adapter approach.  Our method is based on the observation that textual domains are partially overlapping, and we represent domains as a hierarchical tree structure where each node in the tree is associated with a set of adapter weights.  When combined with a frozen pretrained language model, this approach enables parameter sharing among related domains, while avoiding negative interference between unrelated ones. Experimental results with GPT-2 and a large fraction of the 100 most represented websites in C4 show across-the-board improvements in-domain.  We additionally provide an inference time algorithm for a held-out domain and show that averaging over multiple paths through the tree enables further gains in generalization, while adding only a marginal cost to inference.
\end{abstract}

\section{Introduction}
Pretrained language models (PLMs) \cite{peters-etal-2018-deep,  devlin-etal-2019-bert,liu2019roberta,radford2019language}, trained on massive general-domain corpora, have enabled great progress in many natural language processing (NLP) benchmarks \cite{wang-etal-2018-glue}. Nonetheless, continuing pretraining (as a dense model) a PLM on a narrower domain \cite{han-eisenstein-2019-unsupervised, biobert} is beneficial, although computationally expensive \cite{maronikolakis-schutze-2021-multidomain}, which indicates that domain-relevant data is important for downstream tasks. Sparse models that use mixtures of experts \cite{lepikhin2021gshard} have recently been proposed to allow efficient training.

\begin{figure}[!t]
	\centering
	\includegraphics[width=0.95\columnwidth, page=1]{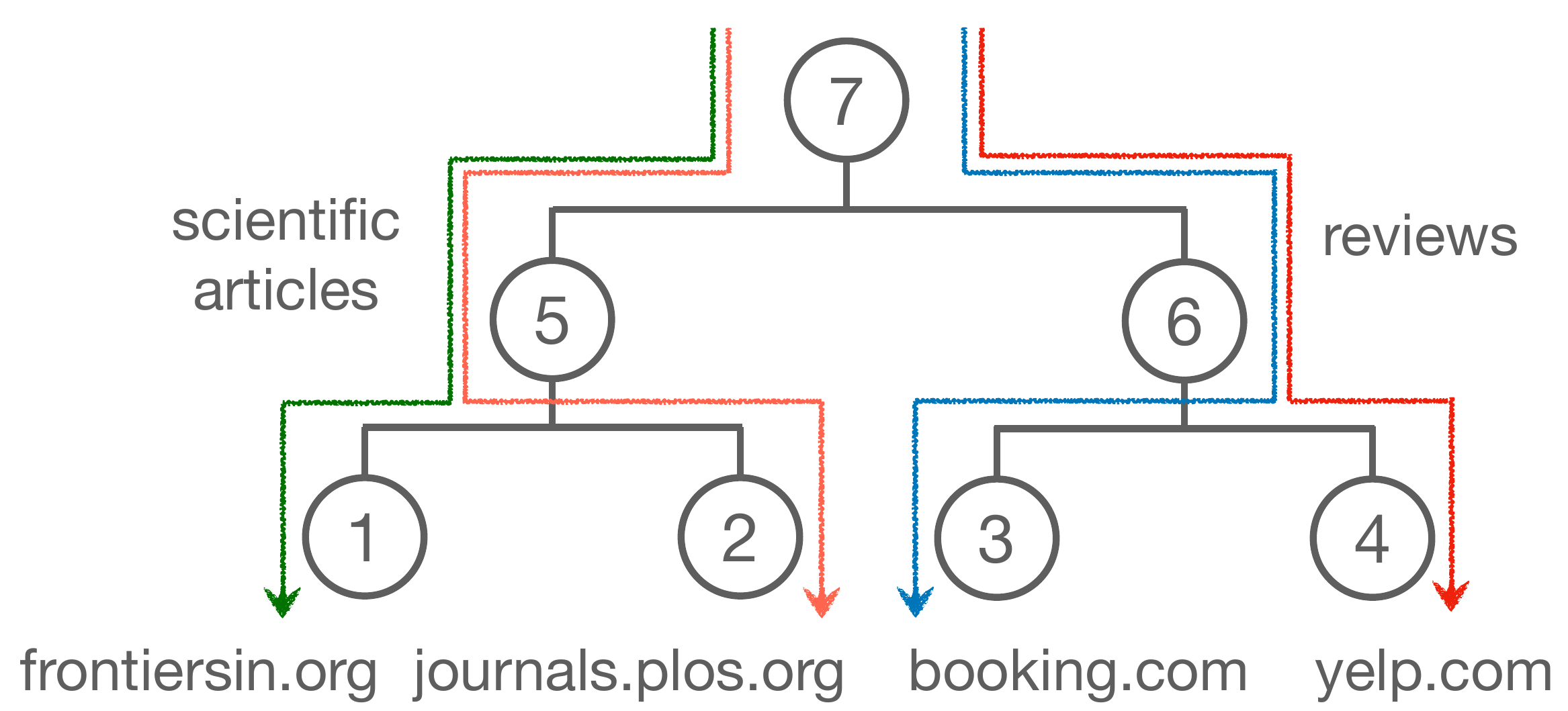}
        	\caption{We model domains as a hierarchical tree structure that associates adapters with nodes, allowing parameter sharing among related domains. Internet domains appear as leaf nodes. During training, we activate the adapters along the path to a leaf to specialize a language model to the domain corresponding to it.
        	}
	\label{fig:hierarchy_example}
\end{figure}

Prior work typically assumes that individual domains are distinct, and models them accordingly. For example, \citet{gururangan-etal-2020-dont,gururangan2021demix} train one model for each textual domain, either in a dense or sparse manner. 
This is related to data selection \cite{moore-lewis-2010-intelligent,axelrod-etal-2011-domain,plank-van-noord-2011-effective}, which aims to select the best matching data for a new domain.
This process does not scale to multiple domains efficiently, as the parameters grow linearly with the domains.  It also does not allow sharing representations among related domains during training, as each domain is modeled with a separate set of parameters.  At the other extreme, training one model on all domains as is common during unsupervised pretraining does not account for their similarities and differences and might hinder the model's generalization ability due to negative interference. 

As an alternative, we start with the observation that the term ``domain'' 
typically denotes a distribution over language characterizing a given topic or genre, and that domains are partially overlapping.  For example, a sentiment model processing hotel reviews could be expected to benefit by also including data from restaurant reviews, which might in turn benefit from cooking recipes, but combing hotel reviews and recipes may be detrimental.

We want to model the relations between domains and selectively share information, so that we allow positive transfer and avoid negative interference. To this end, we propose a data-driven approach to modeling domains that automatically clusters them in a tree using PLM representations. We then introduce an efficient 
method that specializes a PLM in a number of domains leveraging their hierarchical structure. Our approach allows parameter sharing among related domains using adapters \cite{rebuffi, houlsby}, which are lightweight layers, added after each transformer \cite{NIPS2017_3f5ee243} layer. Each node in the tree is associated with a separate set of adapters, that are only activated for a particular domain.  For instance, data from \textsc{booking.com} activates parameters in nodes 3, 6, and 7, allowing parameter sharing with the highly related \textsc{yelp.com} through nodes 6 and 7 (Figure \ref{fig:hierarchy_example}).  

We verify the efficacy of our approach in two settings. First, we manually define a tree structure, using websites
as the leaves.
In this first \textit{few-domain} setting, our method outperforms prior work 
including single  
and multi-domain adapters added to GPT-2 \cite{radford2019language} when tested in-domain. 
We further show that our method generalizes better to held-out websites than the baselines.

We then scale our model to a \textit{many-domain} setting across almost 100 websites.
We induce the hierarchical structure in an unsupervised way using representations from GPT-2 with a Gaussian Mixture Model (GMM) \cite{aharoni-goldberg-2020-unsupervised} and hierarchical clustering, similar to \citet{kldivergence}.
In this way, the clusters model textual domains and the GMM provides a mechanism to automatically find the closest training websites to any held-out website.
Empirical results show across-the-board improvements over strong baselines when evaluated in-domain. 
We also show that an efficient inference-time algorithm that averages over multiple paths through the tree improves generalization when tested on held-out websites.\footnote{Our code is publicly available at \href{https://github.com/alexandra-chron/hierarchical-domain-adaptation}{github.com/alexandra-chron/hierarchical-domain-adaptation}.}

\section{Hierarchical Representation of Domains}

In this section, we provide a formal problem definition and the intuition for a hierarchical ordering of domains. We then  describe how we add a hierarchical structure to a PLM and present the training process. Additionally, we show how a path in the tree is selected to evaluate the in-domain and out-of-domain sets. We finally discuss the computational cost of our approach compared to the baselines and our experimental setup.

\subsection{Problem Definition}
We formulate the task as follows: given a PLM, we aim at fine-tuning it in the task of language modeling using adapter modules, on a corpus consisting of $k$ corpora for domain adaptation. The model is trained to minimize the cross-entropy loss on sentences from all $k$ corpora, then is evaluated on both in-domain and out-of-domain test sets.

\subsection{Textual domains and provenance}
As there is no commonly-accepted definition of a ``domain'' in text \cite{Plank16}, we take a practical approach and use the provenance of a piece of text (that is, the website from which the text was scraped) as a proxy for textual domain.
To model how similar different textual domains are to each other, we fit a Gaussian Mixture Model (GMM) using PLM representations \cite{aharoni-goldberg-2020-unsupervised} using a small sample of text from each domain.
After fitting this GMM, for some clusters there is a one-to-one correspondence between one cluster and text from one website, while for other clusters there is a one-to-many correspondence between one cluster and text from multiple, similar websites (see $\S$\ref{sec:large}).


\subsection{Hierarchical Structure} 

Domains generally overlap with each other and have different degrees of granularity. A model that encodes them should both capture domain-specific and general-domain information.
To this end, we propose representing the data as a tree.
An example of a tree structure is shown in Figure \ref{fig:hierarchy_example}.
Text from specific websites is encoded in the leaf nodes (such as \textsc{frontiersin.org}, \textsc{journals.plos.org}), while more general-domain knowledge is encoded in the upper nodes (\textsc{scientific articles}).

\subsection{Model Architecture}
Assuming a corpus with data from $n$ domains, we consider the setting where we have a pretrained model $M$. We want to use $M$ to adapt to $n$ new domains. To this end, we can leverage \textit{adapters}. 

\noindent \textbf{Adapter layer.} Adapters are typically added to model $M$ in each transformer layer and are trained to a task, while $M$ remains unchanged. An adapter  uses as input the output of the previous layer. 
It is formally defined as $ \mathbf{W}^U \: \text{ReLU}(\mathbf{W}^D \: \text{LN}(h_i)) + h_i$
%
 ,where $h_i$ is the output of the $i$-th layer, of dimension $m$, LN is a layer-normalization \cite{ba2016layer},   $\mathbf{W}^D$ is a down-projection in $R^{m\times d}$, and  $\mathbf{W}^U$ is an up-projection in $R^{d\times m}$, and $d$ is the bottleneck dimension of the adapter module. 

\noindent \textbf{Single Adapters.} 
To adapt to $n$ domains, one solution is to train $n$ adapters (per transformer layer), one for each domain.
The number of parameters added from single adapters grows linearly with the number of domains (O$(n)$). 

 \noindent \textbf{Multi-Domain Adapters.} Another solution is to add just one set of adapters to model $M$. The adapter weights will be updated based on data from all $n$ domains. This is a \textit{dense} model that does not permit modular training. For $n$ domains, the number of parameters added is constant.

\noindent \textbf{Hierarchical Adapters.} 
We propose associating each of the nodes in a tree that represents domains with a set of adapters and adding them to $M$. This \textit{sparse} model adds parameters that scale logarithmically (O$(log(n)$) with the number of domains because of the binary tree structure (Figure \ref{fig:hierarchy_example}). 

While \citet{houlsby} insert adapters but re-train layer normalization parameters of $M$ and \citet{bapna-firat-2019-simple} introduce new layer normalization parameters for every adapter, we introduce just one set of layer normalization parameters in each transformer layer and these parameters are \textit{shared} between all adapters of a transformer layer. 

\subsection{Training \& Computational Cost}
\label{ssec:cost}
 When our input consists of data from a particular domain, we only update the adapter layers of the path that leads to this domain (Figure \ref{fig:hierarchy_example}). 

Supposing we have a mini-batch from \textsc{frontiersin.org}, the hidden state $h_{i}$ of the ${i}$-th layer is the input of $adapter_{i}^1$ (the adapter of node 1 for transformer layer $i$). $h_{i}$ is also the input of $adapter_{i}^5$ (parent) and $adapter_{i}^7$ (root). Their outputs $y_{i}^1, y_{i}^5, y_{i}^7$ are averaged. The final representation $y_{i}$ is the input to the next transformer layer. 

Using this simple training process, we allow sharing between related domains. Upper nodes in the tree are updated more often than leaves, thus they are better trained and encode more domain-general knowledge.
More precisely, the root node of the hierarchical model in Figure \ref{fig:hierarchy_example} is updated for each sequence, but the leaf nodes are only updated using sequences from the associated domain.


In terms of computation, although our model adds a large number of trainable parameters (\textit{total parameters}), only a small fraction of them is used for each forward pass (\textit{active parameters}), as shown in Table \ref{table:parameters}. 
 At inference time, to evaluate performance on a domain using the tree of Figure \ref{fig:hierarchy_example}, our approach with a single path uses 126M parameters (GPT-2 has 112M and the adapters of each path account for 14M parameters). When we average two paths, 23M parameters are added to GPT-2.
 
 \citet{Kaplan2020ScalingLF} provided a detailed breakdown of compute cost for transformer LMs.  For a model with $N$ non-embedding parameters, the approximate cost of a forward pass is $2 N$ flops per token.
 Extending their calculations to our setting, for a model with $L$ layers, model dimension $d_\textrm{model}$, adapter bottleneck size $d$, a single adapter adds $4 L d_{\textrm{model}} d$ flops per forward pass over the cost of running GPT-2.  Our hierarchical method requires running $T$ adapters per layer per forward pass, where $T$ is the average tree depth.  For the many-domain setting in $\S$\ref{sec:large} with $L=12$, $d_{\textrm{model}}=768$, $d=64$, $N=84$M, $T=8$, this gives an increase of  \char`\~ 11\% flops over GPT-2.
 At inference time, using two paths ($\S$\ref{ssec:out_of_domain_large}) the increase is 22\% over GPT-2.

 For fair comparison between our method and the baselines, we scale the adapter size so that our proposed model and the multi-domain adapters (the most related baseline) use the same number of flops.  Following the previous paragraph, the adapter sizes in our hierarchical model are smaller by a factor of $1 / T$ then those in the baseline models (Table \ref{table:parameters}).

\begin{table}[t] 
\resizebox{\columnwidth}{!}{%

\begin{tabular}{llrr} 
                 \toprule                                                                &                     & \textbf{Few-Domain} & \textbf{Many-Domain} \\
                                                                              &                     & \textbf{Setup} & \textbf{Setup} \\\midrule
\multirow{5}{*}{\rotatebox[origin=c]{90}{Hierarchical (ours)}}& \textbf{Adapter Size}        & 256         & 64          \\ 
& \textbf{\# Adapters  }       & 7           & 49          \\
  & \textbf{Average path length} & 3           & 8           \\
 & \textbf{Total parameters}    & 33M         & 58M         \\
 & \textbf{Active parameters}   & 14M         & 9.5M        \\ 
 & \textbf{Number of updates - root} & 22K& 11K \\ 
  & \textbf{Number of updates - leaf} &5.5K & 400 \\ 

 \midrule
\multirow{5}{*}{\rotatebox[origin=c]{90}{{Multi-domain}}}             & \textbf{Adapter Size}        & 768         & 512         \\
      & \# \textbf{Adapters}   & 1           & 1           \\
     & \textbf{Average path length} & 1           & 1           \\
    & \textbf{Total parameters}    & 14M         & 9.5M        \\
     & \textbf{Active parameters}   & 14M         & 9.5M \\
      & \textbf{Number of updates} & 22K & 11K\\

     \bottomrule      
\end{tabular}}
	\caption{Parameters used by our approach and the multi-domain adapters. The \textit{few-domain} and \textit{many-domain setup} are explained in $\S$ \ref{sec:small} and $\S$ \ref{sec:large} respectively.}

	\label{table:parameters}
\end{table}
\subsection{In-domain/Out-of-domain Evaluation}
At inference time, we need to define which path should be activated for each domain. When we perform in-domain evaluation, this is straightforward. We always activate the path that leads to the node that is assigned to this specific domain.

For out-of-domain evaluation, we need to find the path that better fits the held-out domain. We can also use multiple paths, as the computational cost is small. We describe in detail how we run out-of-domain evaluation in the following two sections, which present a manually defined ($\S$\ref{sec:small}) and an automatically created hierarchical structure ($\S$\ref{sec:large}).

\subsection{Experimental Setup}
We use GPT-2 (12 transformer layers; hidden size 768) as the pretrained model. GPT-2 has a vocabulary of 50,264 BPE \cite{sennrich-etal-2016-neural} tokens and 112M parameters. Our code is built with PyTorch \cite{pytorch}, using the HuggingFace library \cite{wolf-etal-2020-transformers}. We run all experiments on NVIDIA A100 GPUs with 40GB of RAM. We split our corpora in 800-token sequences. Models are trained with the Adam optimizer \cite{kingma2014adam} with an initial learning rate of $1e^{-3}$ and we accumulate gradients over 2 updates.

\section{Hierarchical Domain Adaptation with a Manually Created Tree}
\label{sec:small}
In this section, we implement the model described in the prior section for a very limited number of domains (\textit{few-domain setup}), to comprehensively examine design choices and verify the performance, before moving to a large-scale setting in $\S$\ref{sec:large}.

\subsection{Data}
We select four websites to be represented by leaf nodes in our tree: two that contain scientific articles (\textsc{frontiersin.org}, \textsc{journals.plos.org}) and two that contain reviews (\textsc{booking.com} and \textsc{yelp.com}). We use text from the released version \cite{dodge-etal-2021-documenting} of C4 \cite{JMLR:v21:20-074}, a web-scale corpus of English data; the first three internet domains are some of the largest sources of text in C4. We also use \textsc{yelp.com}
, a publicly available dataset. Dataset sizes (training/evaluation) are shown in Appendix \ref{ssec:corpus}.

\subsection{Approach}
We use the hierarchical structure shown in Figure \ref{fig:hierarchy_example}, with two leaf nodes representing scientific articles sharing a parent, two leaf nodes representing reviews sharing a parent, and a single grandparent shared by the two parents. This tree structure was manually chosen using domain knowledge. 
We use a pretrained GPT-2 model as our base model, and add one set of adapters per node in the tree (one adapter per transformer layer for each node).
We freeze the weights of GPT-2 and train the adapters on language modeling of the domains of interest.
The training process is explained in detail in $\S$\ref{ssec:cost}.

\subsection{Experimental Setup}

Our hierarchical model adds 7 sets of adapters to GPT-2, one for each node in the tree. 
Each adapter has a bottleneck dimension $d$ of 256. 
For each training step, one path through the tree is active (so, 3 adapters) depending on which domain of text is represented in the current batch (see $\S$\ref{ssec:cost}).
Active nodes are used in the forward pass and updated in the backward pass (during training), while those that are not active are not used in the computation. 

We evaluate two baselines: a \textit{multi-domain adapter}, trained on all in-domain data, and \textit{single adapters}, each trained on data from a different website. 
 We ensure that the hierarchical model uses the same amount of compute for a forward pass as the multi-domain adapter baseline (using $d$ = $768$ and 1 adapter/path). We also train each model to an equal amount of data from each domain. Results are shown after 20 epochs of training (22K steps).
\begin{table}[h]
\centering
\small
\resizebox{\columnwidth}{!}{%

\begin{tabular}{lrrrr}
\toprule
& \textbf{GPT-2} & \textbf{single}  & \textbf{multi}  & \textbf{hierarchical}  \\ 
&  &  \textbf{adapters} & \textbf{adapters} &  \textbf{adapters} \\ \midrule 
frontiersin& 22.2 & 16.1 & 15.8 & \textbf{15.5}\\
journals   & 24.5 & 16.6 & 16.3 & \textbf{15.8} \\
booking     & 29.7 & 9.7 & 9.9 & \textbf{9.2} \\
yelp & 36.2 & 24.3 & 25.3 & \textbf{23.8} \\ \midrule
average & 27.7 & 15.8 & 15.9 & \textbf{15.2} \\
\bottomrule

\end{tabular}}
\caption{In-domain evaluation perplexity for the few-domain setting ($\S$\ref{sec:small}). Hierarchical adapters consistently provide better scores compared to the baselines.}
\label{table:small_in_domain}
\end{table}

\subsection{In-Domain Results}

In-domain evaluation scores are presented in Table \ref{table:small_in_domain}. Our model clearly surpasses the multi-domain adapter baseline in all domains. On average, hierarchical adapters lower the perplexity by $0.7$ compared to multi-domain adapters. Compared to just evaluating GPT-2, our model yields a large improvement, confirming prior work that suggests that further training a PLM in-domain is highly effective. Single adapters perform roughly equivalent to multi-domain adapters in this scenario. 

\begin{table}[t]
\centering
\small
\resizebox{\columnwidth}{!}{%

\begin{tabular}{lrrrr}
\toprule
                  & \textbf{GPT-2} & \textbf{single}   & \textbf{multi}    & \textbf{hierarchical} \\
                  &      & \textbf{adapters} & \textbf{adapters} & \textbf{adapters}     \\ \midrule
ncbi            & 20.5 &   18.2       & 17.6     & \textbf{17.3}         \\
link.springer    & 27.7 & 24.5         & \textbf{22.7}     & \textbf{22.6}         \\
scholars.duke & 22.7 &  20.1        & 20.3     & \textbf{19.9}    \\
techcrunch        & 27.7 & 27.1         & \textbf{26.3}     & 27.1         \\
medium            & 29.1 & 30.0         & \textbf{27.9}     & 28.5         \\
tripadvisor      & 41.3 & 36.6         & 34.1     & \textbf{26.0}         \\

lonelyplanet      & 35.5 & 27.1         & \textbf{24.3}     & 25.3         \\ \midrule
average           & 29.2 & 26.2         & 24.8     & \textbf{23.8}        \\ \bottomrule
\end{tabular}}
\caption{Out-of-domain evaluation perplexity for the small setting ($\S$ \ref{sec:small}). For the hierarchical model, 2 paths through the tree are used for the evaluation. The hierarchical model on average outperforms the baselines.}
\label{table:small_out_of_domain}
\end{table}

\subsection{Out-of-Domain Results}
We perform evaluation on 7 unseen domains, some of which represent similar textual domains to our in-domain data, while others are quite different. For example, \textsc{ncbi}, \textsc{link.springer}, and \textsc{scholars.duke} contain text from scientific documents, similar to two of our in-domain sources of text, but \textsc{techcrunch} and \textsc{medium} are quite dissimilar to the in-domain text. All models outperform the baseline of just evaluating GPT-2, as shown in Table \ref{table:small_out_of_domain}.
We hypothesize that the pretraining data from GPT-2, which has not been publicly released, had a somewhat different distribution to C4, and thus further training on any data from C4 seems to improve performance. The best out-of-domain results are obtained with hierarchical adapters. 

However, which set of single adapters we should use to evaluate a held-out domain is not obvious. For example, to evaluate on \textsc{lonelyplanet}, it intuitively makes sense to use adapters trained on a reviews/travelling domain  (\textsc{booking} or \textsc{yelp}), but for \textsc{link.springer}, the model trained on a scientific articles (\textsc{frontiersin} or \textsc{journals}) might be more suitable. We have no \textit{a priori} criterion to choose the most appropriate model. This is also true for our proposed model. We show the best evaluation scores using single adapters in Table \ref{table:small_out_of_domain} (full evaluation in Appendix \ref{ssec:fewdomain-appendix}). 

For the hierarchical adapter model, we show evaluation scores using various paths in Table \ref{table:ablation}.  As expected, using a \textit{single} path, the hierarchical model performs best leveraging the path of a website that is most similar to the unseen website. For example, the best evaluation score for \textsc{ncbi} is obtained with the path that leads to \textsc{journals}, while the best score for \textsc{tripadvisor} using the path that leads to \textsc{booking}. Using two paths (either the paths of \textsc{frontiers} and \textsc{journals}, or \textsc{booking} and \textsc{yelp}), results generally improve. For science or technology websites, using the paths of the science domain considerably boosts the hierarchical model's performance. For reviews/travelling websites, using both paths of the reviews domain is beneficial. This confirms our intuition that the hierarchical structure proposed adequately models domains, preventing negative transfer.
\begin{table}[t]
\resizebox{\columnwidth}{!}{%

\begin{tabular}{lrrrr|rr} \toprule
              & \multicolumn{4}{c|}{\textbf{1 path}}   & \multicolumn{2}{c}{\textbf{2 paths}}     \\
   & \textbf{journals} & \textbf{frontiers}&  \textbf{booking} & \textbf{yelp} &\textbf{science} & \textbf{reviews} \\       \midrule 
ncbi  &  \textbf{17.6}& 18.7  & 34.8  & 26.0 & \textbf{17.3} & 26.3 \\
link.springer & \textbf{23.3} &  \textbf{23.3}  &  37.0  &33.1 & \textbf{22.6} & 31.8  \\
scholars.duke & \textbf{20.7} &  \textbf{20.7} & 35.5 &29.4 & \textbf{19.9} &  28.8         \\  
techcrunch  & \textbf{27.7} &27.9 & 34.8  &  32.8  & \textbf{27.1} & 29.4 \\
medium& \textbf{29.4} &  \textbf{29.4} & 35.9 &36.2 &  \textbf{28.5} & 30.6 \\
tripadvisor  & 47.9  & 47.9 & \textbf{37.0}  & 38.1 &   45.6 &  \textbf{26.0}\\
lonelyplanet &39.6 & 40.0& \textbf{25.5}  &  38.9 & 38.5 & \textbf{25.3} \\
      \midrule   average & \textbf{29.5}  &29.7 & 34.4& 33.5  & 28.5 & \textbf{28.3} \\ 
\bottomrule                                       
\end{tabular}}
	\caption{Out-of-domain evaluation of the hierarchical model using different paths. The left part of the table shows scores using a single path. The right part shows results using the average of two paths, corresponding to either the \textit{scientific articles} or the \textit{reviews} domain.}
	\label{table:ablation}
\end{table}

Comparing our hierarchical adapters to multi-domain adapters, using a \textit{single path}, hierarchical adapters perform worse than multi-domain adapters (average scores of columns 1-4 in Table \ref{table:ablation} are worse than the average score of column 3 in Table \ref{table:small_out_of_domain}). 
However, with a \textit{second path} active, hierarchical adapters outperform all other approaches (Table \ref{table:small_out_of_domain}).
This highlights an advantage: they are extensible even after training, allowing for flexible performance-efficiency trade-offs that dense approaches (like multi-domain adapters) do not.

\begin{figure*}[t]
	\centering
	\includegraphics[width=1\textwidth, page=1]{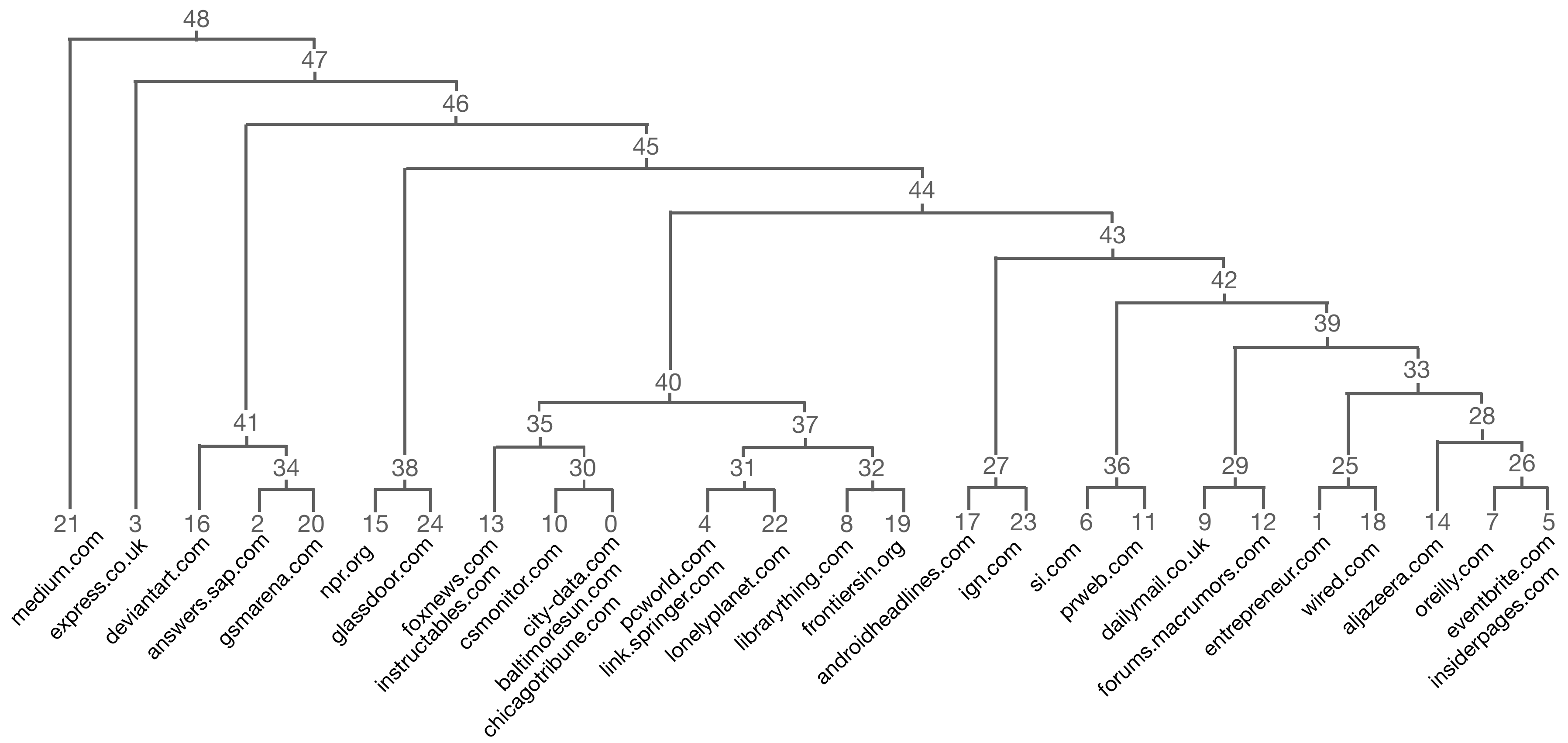}
	\caption{Dendrogram obtained from agglomerative clustering based on the average KL divergences of the GMMs. This diagram illustrates the hierarchical structure of 30 of the most high-resource websites on C4. The leaf nodes correspond to the cluster centers and are mapped to the websites they assign the highest probability to.}
	\label{fig:dendrogram}
\end{figure*}

\section{Hierarchical Domain Adaptation with an Automatically Created Tree}
\label{sec:large}

In this section, we scale our approach to a \textit{many-domain setup}, using a larger set of domains, and thus a much larger hierarchy, adding more adapters in our model. 
In the previous section, we manually selected a tree based on our domain knowledge, but in this section we automatically create a tree using unsupervised methods. We leverage domain clusters obtained using Gaussian Mixture Models and hierarchical clustering and provide an algorithm for out-of-domain evaluation, leveraging the flexibility of hierarchical adapters, that can be combined to improve performance with a minimal cost. 

\subsection{Data}
As a training and evaluation corpus, we use data from C4. Specifically, we use text from 30 websites as our training corpus and we perform out-of-domain evaluation of our model and the baselines on 38 other websites. All websites used belong to the top 100 sites in C4 (details in Appendix \ref{ssec:corpus}).

\subsection{Approach}
We want to create a hierarchical structure that represents relations between domains. To this end, we fit a Gaussian Mixture Model (GMM) and then use an agglomerative clustering algorithm on the GMM. A GMM assumes that all data points are generated from a mixture of a  $k$ Gaussian distributions and defines the probability for data points to belong to any of these distributions. We consider a GMM to be suitable choice because it accounts for the uncertainty of cluster assignment and provides soft assignments that we use at inference.

Similar to \citet{aharoni-goldberg-2020-unsupervised}, we generate contextual representations of 1K  sequences (uniformly sampled) from each of our 30 training websites using GPT-2. We use PCA for dimensionality reduction. We then fit a GMM with 30 components to our data (i.e., 30 Gaussians/clusters). After that, we find the Gaussian which assigns highest probability to text from each website, and remove any Gaussian which does not assign the highest probability to any website (it can be the case that text from more than one websites could be drawn by the same Gaussian). The websites and their corresponding clusters are shown in Figure \ref{fig:dendrogram}.

For hierarchical clustering, we use the symmetrized Kullback-Leibler (KL) divergence as a distance metric. Suppose we have two multivariate normal distributions (means $\mu_0, \mu_1$, covariance matrices $\Sigma_0, \Sigma_1$) obtained by the GMM. To measure the difference between the two distributions, if they have the same dimension $N$, we compute the KL divergence. Because it is asymmetric, we cannot use it to measure the distance between distributions, so we compute the symmetrized version as follows:
\begin{figure*}[t]
	\centering
	\includegraphics[width=1\textwidth, page=1]{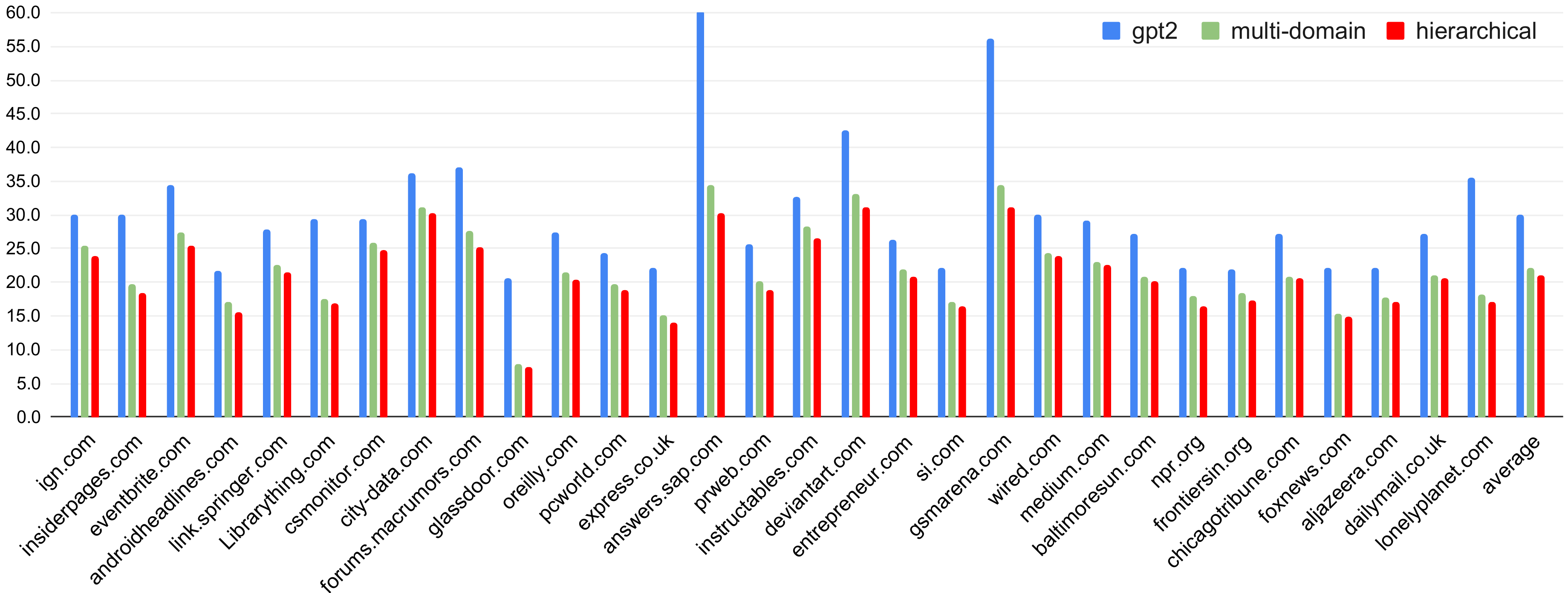}
	\caption{In-domain evaluation perplexity. Hierarchical adapters  consistently outperform the multi-domain adapter on all websites used during training. }
	\label{fig:in_dom}
\end{figure*}
\setlength{\abovedisplayskip}{0pt} \setlength{\abovedisplayshortskip}{0pt}

\scalebox{0.9}{\parbox{.9\linewidth}{%
\begin{align} \label{align:1}
 &D_{KL}(\mathcal{N}_{0}   \| \mathcal{N}_{1}) = \frac{1}{2}  \text{tr} \left(\Sigma_{1}^{-1}\Sigma_{0}) + \ln \left(\frac{\det\Sigma_{1}}{\det \Sigma_{0}}\right)\right)  \nonumber \\
 &+ \frac{1}{2} \left((\mu_1 - \mu_0)^{\text{T}} \Sigma_{1}^{-1}(\mu_1 - \mu_0) -N \right) 
\end{align}
}
}
\scalebox{0.9}{\parbox{.9\linewidth}{%

\begin{align} \label{align:2}
 &D_{KLsym}(\mathcal{N}_{0}, \mathcal{N}_{1}) = \nonumber 
 \\  &\frac{1}{2} \left( D_{KL}(\mathcal{N}_{0}   \| \mathcal{N}_{1}) + D_{KL}(\mathcal{N}_{1}   \| \mathcal{N}_{0}) \right)
\end{align}}}

Using Equation \ref{align:2} as a distance metric, we use agglomerative clustering to infer the structure of our data. We start from 25 clusters, computed by the GMM (5 are ignored because  do not assign a high probability to data samples from any website, see Appendix \ref{ssec:confusion} for the confusion matrix). The clustering algorithm leads to a tree (see Figure \ref{fig:dendrogram}). Nodes 0-24 correspond to the clusters of the GMM. Each website is assigned to a specific cluster.

\subsection{Experimental Setup}
For PCA, we use 100 dimensions. For the hierarchical clustering, we use distances computed using the symmetrized KL divergence. 
We get a tree of 49 nodes, shown in Figure \ref{fig:dendrogram}. We add 49 adapters to GPT-2, one for each node. For a single training step, just one path in the tree is active (as in $\S$\ref{sec:small}).

In this set of experiments, we used our computational budget to compare against our strongest baseline, multi-domain adapters, as that provided the most competitive results in $\S$\ref{sec:small}. Comparing against single adapters could be relevant but we focus on our strongest baseline, as single adapters have shown to be less able to generalize to held-out domains. We train both our hierarchical model and the multi-domain adapter baseline for 4 epochs (11K steps), using 1 GPU per model and stopping after 51 hours. We oversample the low-resource domains to avoid overfitting. We use $d$ = $64$ for hierarchical adapters, as the average path length is 8 and $d$ = $512$ for the multi-domain adapter, since it adds just 1 adapter/transformer layer (Table \ref{table:parameters}).

\subsection{In-Domain Results}

Our in-domain results are shown in Figure \ref{fig:in_dom}. 
To evaluate our model in-domain, we use the path that leads to the cluster that assigns the highest probability to the domain of interest (the same as during training). For example, to evaluate the performance of the model on \textsc{pcworld}, we use the path that leads to cluster 4. The average path length in the tree is 8, so we ``activate'' 8 adapters on average at every training step and also for in-domain evaluation. 
Our approach consistently outperforms multi-domain adapters, yielding +$1.3$ on average in terms of perplexity (see Appendix \ref{ssec:in_dom_larg} for details).

\subsection{Out-of-Domain Results}
\label{ssec:out_of_domain_large}
We perform out-of-domain evaluation on 38 held-out websites (dataset sizes in Appendix \ref{ssec:corpus}). 
We want to automatically find the best path in the tree for a held-out website. To this end, we use the fitted GMM to assign probabilities to data from the held-out websites. We intuitively want to place a held-out website close to similar training websites, so that it can benefit from positive transfer.

To do that, for a given out-of-domain website $i$, we assume we have a set of $N$ sequences (in our experiments $N=1,000$) that we can use to find the best path; this path is used to evaluate the rest of the data from this website (e.g., for computing perplexity). Following a similar procedure to our training regime, we use GPT-2 to encode $N$ sequences, then use the fitted GMM to find the probability assigned to each of the $N$ vectors by each cluster (i.e., each leaf node). The single best path leads to the leaf node that corresponds to cluster $m$, where $m$ assigns the highest probability to the largest fraction of the $N$ sequences from website $i$. The second best path through the tree leads to cluster $n$ that assigns highest probability to the second-most number of the $N$ sequences from website $i$.
Thus, using the GMM clusters and the hierarchical structure, \textit{without training more parameters}, we are able to evaluate out-of-domain data using the adapters that were trained on the most related domains.
This is similar to the ``cached'' setting in \citet{gururangan2021demix}, and it does require a held-out set of $N$ sequences that are only used for finding the best path through the tree (and not for computing perplexity).
This is realistic setting when one has a significant amount of data from a single source, and we leave other approaches (e.g., finding the best path for every input sequence individually) to future work.

We show in Table \ref{table:out_of_dom_large} results of the out-of-domain evaluations. Our hierarchical adapter model outperforms the baseline of just evaluating GPT-2. We notice that using a single path, our approach provides worse results compared to multi-domain adapters.
In this evaluation, the multi-domain adapters and the hierarchical model have the same number of active parameters, but the adapters in the hierarchical model are trained on less data (except the adapter associated with the root,  which has the same number of updates as the multi-domain adapter but is significantly smaller).
However, by having two paths through the tree active, the hierarchical adapter model leverages its modularity and surpasses multi-domain adapters, obtaining an improvement of +$0.6$ in terms of perplexity. 

At inference time, our approach with a single active path uses 122M parameters (112M of GPT-2 and \char`\~ 10M parameters for a path of average length).
When two paths are active, at most 132M parameters are used.
The overhead is thus quite small; if the two paths have some overlap, this computation is potentially significantly less. On average, active parameters in our model are trained on less data than multi-domain adapters (e.g., leaf nodes only see on average 400 updates, as shown in Table \ref{table:parameters}).

\begin{table}[h]
\resizebox{\columnwidth}{!}{

\begin{tabular}{lrrrr}\toprule
\textbf{Out-of-domain} & \textbf{GPT-2} & \textbf{multi}    & {\textbf{hierarchy}} & \textbf{hierarchy}\\
           \textbf{scores}          &      & \textbf{adapters} & \textbf{1 path} &\textbf{2 paths} \\ \midrule
reuters.com          & 20.9 & \textbf{16.0}     & 16.4                  & 16.3              \\
ibtimes.co.uk        & 24.3 & \textbf{19.5}     & 19.7                  & \textbf{19.5}              \\
bbc.com              & 23.6 & 19.1     & 18.9                  & \textbf{18.7}              \\
tripadvisor.com  & 40.4 & 34.8     & 35.9                  & \textbf{33.8}              \\
cnet.com             & 26.8 & 23.3     & \textbf{22.2}                  & 22.9              \\
telegraph.co.uk      & 30.9 & 23.6     & 24.5                  & \textbf{22.2}             \\
theatlantic.com      & 28.5 & \textbf{23.6}     & 23.8                  & \textbf{23.6}              \\
foxbusiness.com      & 22.9 & \textbf{17.5}     & 19.9                  & 18.2              \\
thesun.co.uk         & 26.8 & 19.9     & 19.9                  & \textbf{18.2}              \\
nydailynews.com      & 24.5 & 19.3     & 19.5                  & \textbf{18.2}              \\
dailystar.co.uk      & 20.7 & 13.9     & \textbf{12.2}                  & \textbf{12.2}              \\
fastcompany.com      & 27.9 & 21.3     & 21.5                  & \textbf{20.9}              \\
nypost.com           & 26.3 & 18.9     & 18.9                  & \textbf{18.7}              \\
businessinsider.com  & 24.3 & \textbf{20.5}     & 20.7                  & 20.9              \\
deadline.com         & 33.1 & \textbf{26.3}     & 33.1                  & 26.8              \\
breitbart.com        & 22.9 & \textbf{16.9}     & 17.8                  & 17.1              \\
techcrunch.com       & 27.7 & 21.5     & 21.8                  & \textbf{20.1}              \\
nme.com              & 28.2 & \textbf{20.1}     & 23.8                  & 20.5              \\
fool.com             & 23.8 & \textbf{22.2}     & 22.4                  & \textbf{22.2}              \\
finance.yahoo.com    & 22.6 & \textbf{20.1}     & 20.3                  & \textbf{20.1}              \\
youtube.com          & 15.3 & 14.2     & 14.4                  & \textbf{13.5}              \\
ncbi.nlm.nih.gov     & 20.7 & 18.5     & 18.4                  & \textbf{18.2}              \\
scholars.duke.edu    & 22.6 & 20.7     & \textbf{20.3}                  & \textbf{20.3}              \\
inquisitr.com        & 22.4 & 17.5     & \textbf{16.4}                  & \textbf{16.4}              \\
simple.wikipedia.org & 22.2 & \textbf{19.5}     & 20.5                  & \textbf{19.5}              \\
kickstarter.com      & 26.6 & 24.0     & 24.8                  & \textbf{22.2}              \\
mashable.com         & 27.1 & 22.0     & 22.0                  & \textbf{21.8}              \\
booking.com          & 29.7 & 22.9     & 24.5                  & \textbf{22.0}              \\
etsy.com             & 28.8 & 26.3     & 26.8                  & \textbf{24.5}              \\
fineartamerica.com   & 25.5 & 26.6     & 26.6                  & \textbf{24.5}              \\
github.com           & 32.8 & \textbf{30.3}     & 30.6                  & 30.6              \\
journals.plos.org    & 23.3 & 20.1     & 20.1                  & \textbf{18.2}              \\
itunes.apple.com     & 34.8 & \textbf{28.8}     & 33.1                  & 30.0              \\
agreatertown.com     & 44.7 & 40.0     & 39.6                  & \textbf{35.9}              \\
premium.wpmudev.org  & 31.5 & \textbf{27.7}     & 30.0                  & \textbf{27.7}              \\
homestars.com        & 34.1 & 29.4     & \textbf{28.2}                  & \textbf{28.2}              \\
reference.com        & 28.5 & \textbf{24.5}     & 25.3                  & \textbf{24.5}              \\
cnbc.com             & 21.1 & \textbf{17.6}     & 18.4                  & \textbf{17.6}             \\ \midrule
average              & 26.8 & \textbf{22.3}     & 23.0                  & \textbf{21.7}               \\ \bottomrule  
\end{tabular}}
\caption{Out-of-domain evaluation perplexity. With 1 path, our hierarchical model performs worse than the baseline. However, using paths of the 2 closest clusters to a held-out website, our approach yields better results. We show the paths used in detail in Appendix \ref{ssec:confusion}.  }
\label{table:out_of_dom_large}
\end{table}

\section{Related work}
Our approach draws on prior work in domain adaption and efficient language model fine-tuning.

\noindent \textbf{Domain Adaptation. } 
A large research area in NLP is domain adaptation \cite{jiang-zhai-2007-instance, daume-iii-2007-frustratingly}. Training a masked language model on data from a specific domain \cite{lee2020biobert, beltagy-etal-2019-scibert} or fine-tuning a PLM using data from the target task \cite{howard-ruder-2018-universal} or the target domain \cite{rietzler-etal-2020-adapt, han-eisenstein-2019-unsupervised} has shown to be helpful to mitigate the domain shift between train and test data distributions of the same task. \citet{gururangan-etal-2020-dont} showed that a PLM can further improve by fine-tuning on data from a domain that is related to the domain of the task (DAPT).  While this work suggests fine-tuning a different model to the domain of each task, our approach trains a single model to adapt to all domains. Also, although DAPT does not permit parameter sharing between domains, our hierarchical adapter model leverages domain similarities to improve adaptation.

Domain expert mixture (DEMix) layers \cite{gururangan2021demix} that condition a LM on the domain of input text have been recently proposed. DEMix layers replace feed-forward layers in a transformer and each of them is updated only using data from a specific domain. Then, a modular LM is trained from scratch. On the contrary, we use a PLM and only train adapter layers on the target domains. 
Since each feed-forward layer is replaced with a mixture of experts, the parameters added grow linearly with the  domains. In our approach, however, the number of parameters grows logarithmically, due to the hierarchical structure. 

\noindent \textbf{Adapters.} Efficient fine-tuning using adapters \cite{rebuffi, houlsby} is prevalent in many NLP taks, such as machine translation \cite{bapna-firat-2019-simple}, cross-lingual transfer \cite{pfeiffer-etal-2020-mad} and dependency parsing \cite{ustun-etal-2020-udapter}. Adapters can be trained on a single task or language \cite{pfeiffer-etal-2020-mad}, but also multilingually  \cite{cooper-stickland-etal-2021-recipes}. They have also been used to infuse factual and linguistic  \cite{wang-etal-2021-k} as well as general-purpose and commonsense knowledge \cite{lauscher-etal-2020-common} into a PLM. To the best of our knowledge, we are the first to use them in a hierarchical structure for domain adaptation. 

\noindent \textbf{Efficient fine-tuning methods for PLMs.} Besides adapters, multiple other parameter-efficient methods to adapt general-purpose PLMs to specific tasks have been recently proposed. Prefix tuning \cite{li-liang-2021-prefix}, low-rank matrix approximation \cite{hu2022lora}, as well as fine-tuning only the bias terms of a PLM \cite{bitfit} are some of the lightweight alternatives to fine-tuning the entire PLM. \citet{he2022towards} show that these fine-tuning methods can be seen as modifications to some specific hidden states of PLMs and can thus be recast. We use adapters in this work, but our method could possibly also benefit from other parameter-efficient approaches. 

\section{Conclusion \& Future Work}
In this paper, we present a novel approach for efficient domain adaptation on multiple domains using hierarchical adapters that encode the similarities and differences of domains, allowing  parameter sharing but avoiding negative transfer. We start with a manually defined tree and then scale to a large tree, created in an unsupervised way. We also provide an evaluation-time algorithm that can combine paths to best adapt to an unseen domain. 

In the future, we would like to investigate a more efficient evaluation-time approach, using only a few tokens of an unseen domain. 
It would also be interesting to extend our model to a multi-lingual setup. Finally, we would like to use our method to control language generation of PLMs, in order to avoid generating hate speech or toxic text. 

\section{Limitations and Risks}

Our work uses generative pretrained language models. As such models are trained on large datasets from text in the Internet, they encode biases that could harm marginalized populations \cite{stochastic-parrots}. The specialized language model we propose could be used for propaganda or hate speech generation, same as any other language model. However, our hierarchical adapter model permits adding modular components and we believe that it could potentially be used to detoxify language generation, following \citet{liu-etal-2021-dexperts}. This is in line with recent work on sparse models  \cite{gururangan2021demix, artetxe2021efficient}. 

\section*{Acknowledgements}
We thank the AllenNLP team and other researchers at the Allen Institute for AI for their thoughtful comments. We also thank Alex Fraser, Dario Stojanovski, and Suchin Gururangan for helpful discussions about this work.

\bibliography{custom}
\bibliographystyle{acl_natbib}

\newpage
\clearpage
\appendix

\section{Appendix}
\label{sec:appendix}

\subsection{Corpus description}
\label{ssec:corpus}
In Table \ref{table:corpora}, we present the sizes of the training and evaluation corpora used for the many-domain setup. Only one corpus is used for the few-domain but not the many-domain experimental setting, namely \textsc{yelp.com}\footnote{\href{https://www.yelp.com/dataset}{www.yelp.com/dataset}}. This corpus has 684M training tokens and 20M evaluation tokens. We randomly sub-sample 53M training tokens of this corpus for our first, few-domain setup, as we want to train a balanced model. Extensive documentation of the corpus is available from  \citet{dodge-etal-2021-documenting}. We use the C4 corpus in accordance with the terms of use\footnote{\href{https://commoncrawl.org/terms-of-use/}{commoncrawl.org/terms-of-use/}}.

\subsection{Few-Domain Setup}
\label{ssec:fewdomain-appendix}
In Table \ref{table:small_out_of_domain}, we present out-of-domain evaluation perplexities for the first experimental setup. For the \textit{single adapters} model, we present the best-performing models in the table. In order to allow for an exhaustive comparison, we also present the evaluation results of all the trained single adapter models on all held-out websites in Table \ref{table:ood_small_single_ablation}. We see, for example, that when evaluating on the traveling website \textsc{tripadvisor}, the single adapter model that was trained on either \textsc{booking} or \textsc{yelp} provides the lowest perplexity scores, confirming our intuition that for a held-out website, we should use a model trained on a very similar domain.

\begin{table}[h]
\centering
\small
\resizebox{\columnwidth}{!}{%

\begin{tabular}{lrrrr|r}
\toprule
& \multicolumn{4}{c|}{\textbf{single adapters trained on}} & \textbf{baseline} \\ 
                  & \textbf{booking} & \textbf{yelp}   & \textbf{frontiers}    & \textbf{journals} & \textbf{GPT-2} \\ \midrule
ncbi            & 20.1 & 20.1 & 19.7 & \underline{18.2} & 20.5     \\
link.springer   & 27.1 & 27.1 & \underline{24.5} & \underline{24.5} & 27.7  \\
scholars.duke   & 22.2 & 22.2 & 22.2 & \underline{20.1} & 22.7 \\
techcrunch      & \underline{27.1} & \underline{27.1} & \underline{27.1} & \underline{27.1} & 27.7 \\
medium          & \underline{30.0} & \underline{30.0} & \underline{30.0} & \underline{30.0} & 29.1 \\
tripadvisor     & \underline{36.6} & \underline{36.6} & 49.4 & 49.4 & 41.3 \\
lonelyplanet    & 30.0 & \underline{27.1} & 40.4 & 40.4 & 35.5  \\ \bottomrule
\end{tabular}}
\caption{Out-of-domain evaluation of single adapters in the few-domain setup ($\S$\ref{sec:small}). We evaluate every set of single adapters in 7 different websites. The best results for every out-of-domain website (underlined) are shown in Table \ref{table:small_out_of_domain}.}
\label{table:ood_small_single_ablation}
\end{table}
\begin{table}[t]
\centering
\small
\resizebox{\columnwidth}{!}{%

\begin{tabular}{lrrr}
\toprule
\textbf{In-domain scores}         & \textbf{GPT-2} & \textbf{multi}    & \textbf{hierarchical} \\
                         &      & \textbf{adapters} & \textbf{adapters}     \\ \midrule
ign.com              & 30.0 & 25.5     & 23.8         \\
insiderpages.com         & 30.0 & 19.7     & 18.4         \\
eventbrite.com       & 34.5 & 27.4     & 25.5         \\
androidheadlines.com & 21.8 & 17.1     & 16.0         \\
link.springer.com        & 27.9 & 22.6     & 21.5         \\
librarything.com         & 29.4 & 17.6     & 16.9         \\
csmonitor.com            & 29.4 & 25.8     & 24.8         \\
city-data.com            & 36.2 & 31.2     & 30.3         \\
forums.macrumors.com     & 37.0 & 27.7     & 26.0         \\
glassdoor.com            & 20.7 & 7.9      & 7.5          \\
oreilly.com              & 27.4 & 21.5     & 20.5         \\
pcworld.com              & 24.3 & 19.7     & 18.9         \\
express.co.uk            & 22.2 & 15.0     & 14.0         \\
answers.sap.com          & 60.3 & 34.5     & 30.3         \\
prweb.com                & 25.8 & 20.1     & 18.9         \\
instructables.com        & 32.8 & 28.2     & 26.6         \\
deviantart.com           & 42.5 & 33.1     & 31.2         \\
entrepreneur.com         & 26.3 & 22.0     & 20.9         \\
si.com                   & 22.2 & 17.3     & 16.4         \\
gsmarena.com             & 56.3 & 34.5     & 31.2         \\
wired.com                & 30.0 & 24.3     & 23.8         \\
medium.com               & 29.1 & 23.1     & 22.6         \\
baltimoresun.com         & 27.1 & 20.9     & 20.1         \\
npr.org                  & 22.2 & 18.0     & 17.5         \\
frontiersin.org          & 22.0 & 18.4     & 17.2         \\
chicagotribune.com       & 27.1 & 21.1     & 20.7         \\
foxnews.com              & 22.2 & 15.3     & 14.9         \\
aljazeera.com            & 22.2 & 17.8     & 17.1         \\
dailymail.co.uk          & 27.1 & 21.1     & 20.7         \\
lonelyplanet.com         & 35.5 & 19.5     & 17.1         \\ \midrule
average                  & 30.0 & \textbf{22.3}     & \textbf{21.0}        \\ \bottomrule
\end{tabular}}
\caption{In-domain evaluation perplexity for the many-domain setup (we note that the hierarchical model uses a single path).}
\label{table:in_dom_large}
\end{table}
\begin{table*}[h]
 \resizebox{0.8\textwidth}{!}{

\begin{tabular}{rrr|rrr} \toprule
  &            & Train (Eval.) Tokens &     &                      & Train (Eval.) Tokens \\ \midrule 
\multirow{37}{*}{\rotatebox[origin=c]{90}{\textsc{Training Corpus}}} & frontiersin.org              & 38M (6M)             & \multirow{37}{*}{\rotatebox[origin=c]{90}{\textsc{Evaluation Corpus}}} & journals.plos.org    & 53M (6M)             \\

                                                                             & chicagotribune.com   & 31M (4M)             &                                                                                & fool.com             & 34M (4M)             \\
                                                                             & link.springer.com        & 28M (4M)             &                                                                                & businessinsider.com  & 32M (4M)             \\
                                                                             & aljazeera.com        & 26M (3M)             &                                                                                & theatlantic.com      & 30M (4M)             \\
                                                                             & instructables.com    & 25M (3M)             &                                                                                & booking.com          & 30M (4M)             \\
                                                                             & npr.org              & 25M (3M)             &                                                                                & kickstarter.com      & 26M (3M)             \\
                                                                             & dailymail.co.uk      & 25M (3M)             &                                                                                & telegraph.co.uk      & 25M (3M)             \\
                                                                             & csmonitor.com        & 23M (3M)             &                                                                                & cnet.com             & 24M (3M)             \\
                                                                             & baltimoresun.com     & 23M (3M)             &                                                                                & ncbi.nlm.nih.gov     & 23M (3M)             \\
                                                                             & city-data.com        & 22M (3M)             &                                                                                & foxbusiness.com      & 23M (3M)             \\
                                                                             & forums.macrumors.com     & 22M (3M)             &                                                                                & cnbc.com             & 20M (2M)             \\
                                                                             & medium.com               & 22M (3M)             &                                                                                & ibtimes.co.uk        & 18M (2M)             \\
                                                                             & foxnews.com          & 22M (3M)             &                                                                                & reuters.com          & 17M (2M)             \\
                                                                             & si.com               & 18M (2M)             &                                                                                & bbc.com              & 17M (2M)             \\
                                                                             & wired.com            & 18M (2M)             &                                                                                & nypost.com           & 15M (2M)             \\
                                                                             & prweb.com            & 17M (2M)             &                                                                                & nydailynews.com      & 14M (2M)             \\
                                                                             & express.co.uk        & 16M (2M)             &                                                                                & fastcompany.com      & 14M (2M)             \\
                                                                             & entrepreneur.com     & 16M (2M)             &                                                                                & mashable.com         & 14M (2M)             \\
                                                                             & androidheadlines.com & 14M (2M)             &                                                                                & thesun.co.uk         & 13M (2M)             \\
                                                                             & pcworld.com          & 14M (2M)             &                                                                                & techcrunch.com       & 13M (2M)             \\
                                                                             & gsmarena.com         & 12M (2M)             &                                                                                & inquisitr.com        & 13M (2M)             \\
                                                                             & eventbrite.com       & 11M (1M)             &                                                                                & youtube.com          & 11M (1M)             \\
                                                                             & ign.com              & 10M (1M)             &                                                                                & itunes.apple.com     & 11M (1M)             \\
                                                                             & oreilly.com          & 9M (1M)              &                                                                                & breitbart.com        & 10M (1M)             \\
                                                                             & deviantart.com       & 9M (1M)              &                                                                                & etsy.com             & 10M (1M)             \\
                                                                             & insiderpages.com     & 8M (1M)              &                                                                                & github.com           & 10M (1M)             \\
                                                                             & lonelyplanet.com         & 6M (1M)              &                                                                                & agreatertown.com     & 9M (1M)              \\
                                                                             & answers.sap.com          & 6M (1M)              &                                                                                & premium.wpmudev.org  & 9M (1M)              \\
                                                                             & glassdoor.com        & 4M (500K)            &                                                                                & deadline.com         & 9M (1M)              \\
                                                                             & librarything.com     & 3M (500K)            &                                                                                & dailystar.co.uk      & 9M (1M)              \\
                                                                             &                          &                      &                                                                                & reference.com        & 7M (1M)              \\
                                                                             &                          &                      &                                                                                & scholars.duke.edu    & 7M (1M)              \\
                                                                             &                          &                      &                                                                                & tripadvisor.com  & 7M (1M)              \\
                                                                             &                          &                      &                                                                                & simple.wikipedia.org & 6M (1M)              \\
                                                                             &                          &                      &                                                                                & nme.com              & 5M (1M)              \\
                                                                             &                          &                      &                                                                                & homestars.com        & 3M (500K)            \\
                                                                             &                          &                      &                                                                                & fineartamerica.com   & 2M (500K)     \\ \bottomrule       
\end{tabular}}
\caption{Domains that make up our in-domain (training) and out of-domain (evaluation) corpus for the large setup, including the size of our training and evaluation data. All data is extracted from C4  \cite{JMLR:v21:20-074}.}
\label{table:corpora}
\end{table*}

\subsection{Many-Domain Setup}
\label{ssec:confusion}

\begin{figure*}[h]
	\centering
	\includegraphics[width=1.\textwidth, page=1]{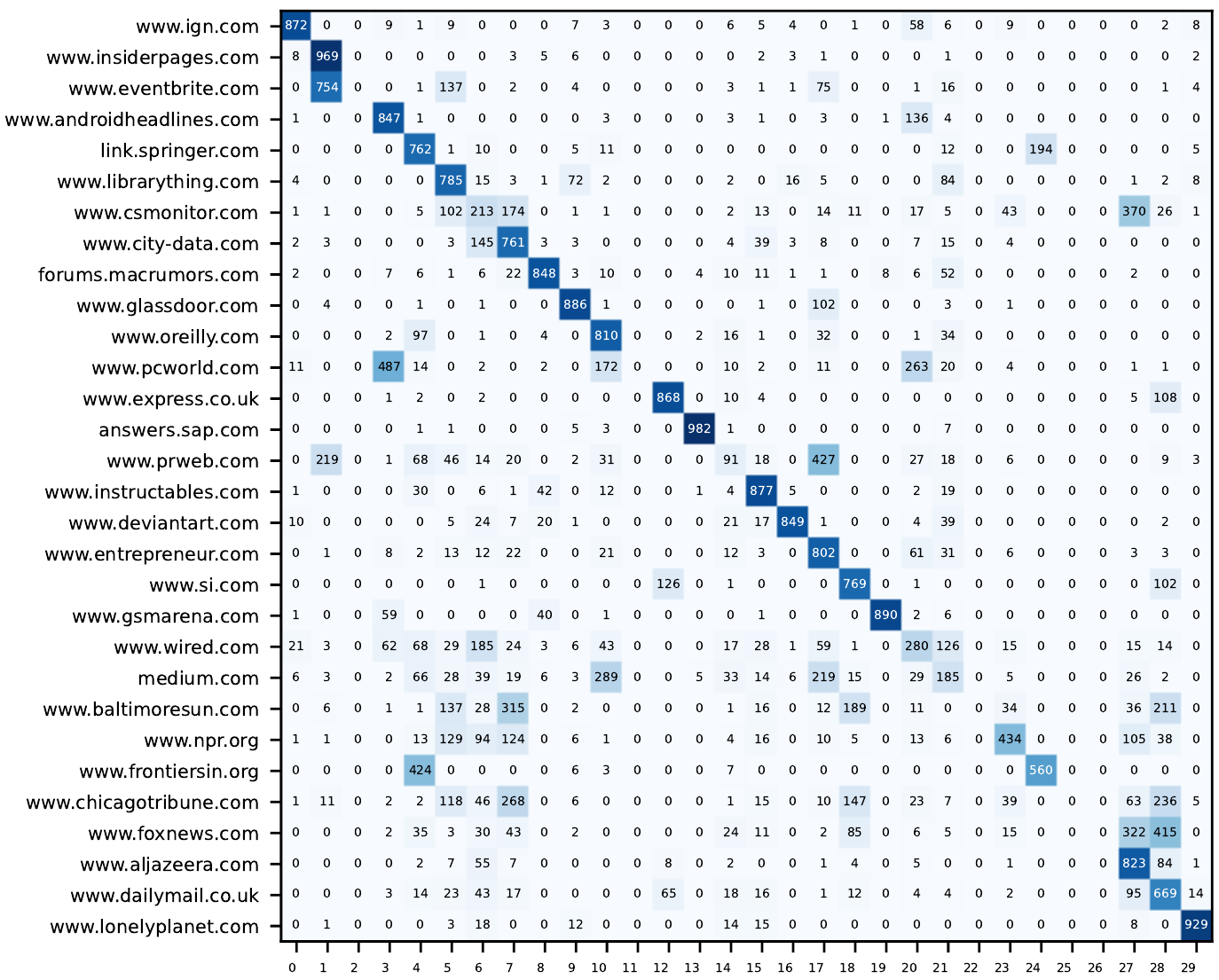}
	\caption{Confusion Matrix. The x-axis depicts the clusters  that the internet domains are assigned to. If no data samples are added to a cluster (for example, cluster 2), the corresponding Gaussian distribution is not used for the hierarchical clustering. The y-axis depicts the internet domains used for training. The cluster numbers shown here are not the exact ones shown in the final dendrogram, but one can easily observe that, for example, the same cluster (in this example, cluster 7) assigns the highest probability to \textsc{city-data.com}, \textsc{baltimoresun.com} and \textsc{chicagotribune.com}. This is mirrored in Figure \ref{fig:dendrogram} of the main paper.}
	\label{fig:confusion}
\end{figure*}

\noindent 
\textbf{Confusion Matrix.} 
Figure \ref{fig:confusion} depicts the confusion matrix of the GMM. We can observe visually that some clusters assign a high probability to multiple internet domains, while others remain empty. This shows that the intuition have for what a domain is does not correspond exactly to the cluster obtained by an unsupervised, data-driven approach. Our visualization is based on publicly available code\footnote{\href{https://github.com/roeeaharoni/unsupervised-domain-clusters}{github.com/roeeaharoni/unsupervised-domain-clusters}}.

\noindent 
\textbf{Out-of-domain Evaluation.} As mentioned in $\S$\ref{ssec:out_of_domain_large}, to run evaluation on a given out-of-domain website $i$, we use two paths of the trained hierarchical model. The first path leads to the leaf node that corresponds to cluster $m$ (with $m$ assigning the highest probability to the largest fraction of $N$ sequences from website $i$) and the second path leads to cluster $n$, where $n$ assigns the highest probability to the second-most number of the $N$ sequences. We present the clusters $m$ and $n$ (and the websites they were mapped to during training) in Table \ref{table:which_path}.

\subsection{Experimental Details}
Because we wanted to keep a modest computational budget, we did not perform multiple training runs for the hierarchical models and the baselines. Results are reported over a single run. 

\begin{table*}[h]
\resizebox{\textwidth}{!}{

\begin{tabular}{lrrrr}\toprule
\textbf{Held-out Website}       &  \textbf{Path to Cluster $m$}   & \textbf{Train. Website of Cluster $m$} & \textbf{Path to Cluster $n$} & \textbf{Train. Website of Cluster $n$} \\ \midrule
reuters.com    &  14  & aljazeera.com  & 15 &   npr.org         \\
ibtimes.co.uk  & 9   & dailymail.co.uk  & 14 & aljazeera.com         \\
bbc.com         & 9  & dailymail.co.uk  & 3 & express.co.uk          \\
tripadvisor.com    & 5  & insiderpages.com & 22 & lonelyplanet.com    \\
cnet.com        & 18  & wired.com   & 17   & androidheadlines.com     \\
telegraph.co.uk & 9  & dailymail.co.uk  & 14  & aljazeera.com       \\
theatlantic.com & 0 & city-data.com & 14  & aljazeera.com    \\
foxbusiness.com & 11 & prweb.com    & 15 & npr.org    \\
thesun.co.uk    & 9 & dailymail.co.uk & 3 & express.co.uk          \\
nydailynews.com & 9 & dailymail.co.uk & 6 & si.com           \\
dailystar.co.uk & 3 & express.co.uk   & 9  & dailymail.co.uk            \\
fastcompany.com & 1 & entrepreneur.com & 18  & wired.com       \\
nypost.com      & 9 & dailymail.co.uk  & 6 & si.com      \\
businessinsider.com & 1 & entrepreneur.com & 18  & wired.com    \\
deadline.com    & 8 & librarything.com     & 9 & dailymail.co.uk          \\
breitbart.com   & 14  & aljazeera.com      & 0   & city-data.com         \\
techcrunch.com  & 18   & wired.com        & 1 & entrepreneur.com       \\
nme.com         & 8  & librarything.com   & 9 & dailymail.co.uk        \\
fool.com        & 1 & entrepreneur.com    & 15  & npr.org   \\
finance.yahoo.com & 1 & entrepreneur.com  & 15  & npr.org     \\
youtube.com     & 11 & prweb.com       & 15  & npr.org    \\
ncbi.nlm.nih.gov & 4 & link.springer.com   & 19 & frontiersin.org         \\
scholars.duke.edu & 4 & link.springer.com & 19  & frontiersin.org      \\
inquisitr.com    & 9  & dailymail.co.uk & 18   & wired.com         \\
simple.wikipedia.org & 10  & csmonitor.com  & 8  & librarything.com    \\
kickstarter.com & 16 & deviantart.com   & 18   & wired.com      \\
mashable.com    & 18  & wired.com    & 9 & dailymail.co.uk    \\
booking.com     & 5 & insiderpages.com & 22 & lonelyplanet.com   \\
etsy.com        & 13 & instructables.com  & 5 & insiderpages.com   \\
fineartamerica.com  & 16 & deviantart.com & 13 & instructables.com   \\
github.com      & 7  & oreilly.com        & 2  & answers.sap.com \\
journals.plos.org & 4 & link.springer.com & 19 & frontiersin.org      \\
itunes.apple.com  & 20 & gsmarena.com      & 8  & librarything.com        \\
agreatertown.com  & 22 & lonelyplanet.com & 5  & insiderpages.com       \\
premium.wpmudev.org & 2  & answers.sap.com & 7    & oreilly.com    \\
homestars.com   & 5 & insiderpages.com   & 13 & instructables.com        \\
reference.com   & 13 & instructables.com & 10 & csmonitor.com        \\
cnbc.com        & 15 & npr.org         & 1 & entrepreneur.com\\ \bottomrule
\end{tabular}}
\caption{The two paths used for evaluation of the hierarchical adapter model on each held-out website.  }
\label{table:which_path}
\end{table*}

\label{ssec:in_dom_larg}

\end{document}